%% file: ms.tex
\documentclass[runningheads]{llncs}
\pdfoutput=1
\usepackage{graphicx}
\usepackage{amsmath,amssymb} %
\usepackage{color}
\usepackage{amsfonts}
\usepackage{bbm}

\DeclareMathOperator*{\argmin}{arg\,min}

\newcommand{\myparagraph}[1]{\vspace{0.1em}\noindent\textbf{#1}}

\newcommand{\bve}{\mathbf e}

\begin{document}
\pagestyle{headings}
\mainmatter

\title{Efficient Full Image Interactive Segmentation by Leveraging Within-image Appearance Similarity}

\author{Mykhaylo Andriluka \and Stefano Pellegrini \and Stefan Popov \and Vittorio Ferrari}

\institute{Google Research}

\titlerunning{Efficient Full Image Interactive Segmentation}
\authorrunning{M. Andriluka \and S. Pellegrini \and S. Popov \and V. Ferrari}
\maketitle

\vspace*{-0.5cm}

\begin{abstract}
We propose a new approach to interactive full-image semantic segmentation which enables quickly collecting training data for new datasets
with previously unseen semantic classes
\footnote{A demo is available at https://youtu.be/yUk8D5gEX-o}.
We leverage a key observation: propagation from labeled to unlabeled pixels does not necessarily
require class-specific knowledge, but can be done purely based on appearance similarity
\textit{within an image}. We build on this observation and propose an approach capable of
jointly propagating pixel labels from multiple classes without having explicit class-specific appearance models.
To enable long-range propagation, our approach first globally measures appearance similarity
between labeled and unlabeled pixels across the entire image. Then it locally integrates per-pixel
measurements which improves the accuracy at boundaries and removes noisy label switches in homogeneous regions.
We also design an efficient manual annotation interface that extends the traditional polygon drawing tools with a suite of additional convenient features (and add automatic propagation to it).
Experiments with human annotators on the COCO Panoptic Challenge dataset show that the combination of our better manual interface and our novel automatic propagation mechanism leads to reducing annotation time by more than factor of $2\times$ compared to polygon drawing.
We also test our method on the ADE-20k and Fashionista datasets without making any dataset-specific adaptation nor retraining our model, demonstrating that it can generalize to new datasets and visual classes.
\end{abstract}

\vspace*{-0.3cm}
\input{intro}

\vspace*{-0.15cm}
\input{related_work}

\vspace*{-0.15cm}
\input{mask_painting}

\vspace*{-0.15cm}
\input{model}

\vspace*{-0.15cm}
\input{experiments}
\vspace*{-0.15cm}
\input{conclusion}
\clearpage
\bibliographystyle{splncs}
\bibliography{shortstrings,loco,additional_refs}

\end{document}

%% file: intro.tex
\section{Introduction}

We consider the task of full-image semantic image segmentation:
assigning a semantic label to each pixel in an image. Existing computer vision models for
this task such as \cite{chen18pami} are powerful, but require a large set of
labeled training images to perform well. However, collecting training data for semantic segmentation
is notoriously time consuming. For example for the popular COCO dataset, the average annotation
time per single foreground object with a polygon outline amounts to 80 seconds \cite{lin14eccv}. Additionally, dense
annotation of the background classes even at the superpixel level required an additional $3$ minutes per
image \cite{caesar18cvpr}. In total, this results in average annotation time of about $19$ minutes per image for the COCO Panoptic dataset \cite{kirillov18coco}.

Several approaches have been proposed to speed-up the annotation of training data based on the interaction between a human annotator and a machine model.
The majority of such methods aim at speeding-up the segmentation of individual objects~\cite{benenson19cvpr,castrejon17cvpr,xu16cvpr,liew17iccv,benard17arxiv,mahadevan18bmvc,li18cvpr,jang2019cvpr,hu19nn,le18eccv,maninis18cvpr,majumder19cvpr,sofiiuk2020cvpr,zheng2020cvpr}.
Applying these approaches to full-image annotation would require breaking down the annotation process into a series of single-object micro-tasks, which is arguably suboptimal as it would require the annotator to parse the image multiple times.
Recently, a few approaches tackle the full-image segmentation annotation task~\cite{andriluka18acmmm,agustsson19cvpr}.
However, they train the segmentation model underlying their interactive system on existing labeled images of {\em the same semantic classes} that appear in the target images to be annotated with it afterwards.
Hence, they are unlikely to work well on datasets with novel classes they have not been trained on, and indeed they have not been demonstrated to do so.

\input{fig_teaser}

In this paper we propose an interactive approach for full-image semantic segmentation annotation that aims to be faster than existing tools.
Our interface builds on the intuition that the visual appearance of an object class {\em within an image} usually has much lower variability than across all images in a dataset.
Hence it is feasible to build a sufficiently good \textit{image-specific} segmentation model on the fly during the annotation process given a few examples provided by the annotator, and then use this model for propagation to previously unlabeled pixels.
We show an example in Fig.~\ref{fig:teaser}. Notice how the annotated pixels for ``person'', ``wall''
and ``grass'' are not sufficient to capture appearance of these classes in general.
However they still allow to correctly classify large portion of image pixels {\em within this image}, enabling effective label propagation (more examples in Fig.~\ref{fig:tools} and \ref{fig:example_results}).

The starting point, and our {\bf first contribution}, is an efficient annotation interface for manual pixel labeling (dubbed `Mask Paint', Sec~\ref{subsec:manualtools}). It combines the strengths of traditional polygon drawing commonly used to annotated segmentation datasets~\cite{lin14eccv,russell08ijcv,zhou17cvpr,cordts16cvpr,mottaghi14cvpr,liu09cvpr,gupta2019cvpr}
and of free-form drawing tools popular in painting programs such as MS Paint and Adobe Photoshop~\cite{photoshop18}.
We demonstrate experimentally that this leads to faster annotation compared to polygon drawing.

Building on this manual interface, our {\bf main contribution} is to augment it with a component that acts as a visual version of ``auto-complete'' for text \cite{witten1992keyboard} (dubbed `Magic Paint', sec.~\ref{subsec:autotools} and~\ref{sec:model}). This component attempts to automatically propagate pixel labels across the image based on the appearance similarity between regions that have been previously manually labeled and regions that have not been annotated yet.
We define both an inference system for label propagation that can deal with a variable number of labels defined by the annotator (Sec.~\ref{sec:model}), as well as new user interface features to effectively use the propagated labels during the annotation process (see Sec.~\ref{subsec:autotools}).
Typically, our label propagation component saves time to the human annotator by filling in image parts similar to previously labeled regions, allowing the annotator to focus on the more difficult parts of the image only.
In Fig.~\ref{fig:teaser} we illustrate how the typical annotation process in our system differs from traditional polygon drawing.
Whereas annotators using polygon drawing interfaces spend most of their effort in labeling object class boundaries. In our system annotators usually start by marking pixels in the interior of the class and let the system infer the boundaries which is more time efficient (as we show in Sec.~\ref{sec:experiments}).
Furthermore it might lead to more accurate boundaries, as the propagator leverages fine-grained pixel gradients to snap to exact boundaries (observe the sharp boundaries between ``person''/``wall'' and ``grass''/''person'' classes in Fig.~\ref{fig:teaser}).
To maximize the effectiveness of the propagation component we allow it to propagate labels anywhere in the image. For this we rely on global appearance matching between the labeled and unlabeled pixels.
We represent appearance with learned embedding vectors which improve the quality of such
matching compared to simpler appearance representations such as color histograms.
Given the output of the global appearance matching, we then locally integrate evidence with a dense conditional random field~\cite{kraehenbuehl11nips}, which allows to smooth-out spurious label switches in the middle of homogeneous regions, and to snap to exact boundaries.

As a {\bf third contribution} we introduce a benchmark for annotation evaluation that includes images from three datasets: COCO Panoptic Challange~\cite{kirillov18coco}, ADE 20k~\cite{zhou17cvpr}, and Fashionista~\cite{yamaguchi12cvpr}.
We provide recordings of real user annotation sessions and an evaluation protocol that enables us to assess the performance of interactive segmentation using simulation based on the real annotator input.
Our benchmark allows us to go beyond simulation based purely on sampling artificial point click inputs as is common in the literature \cite{mahadevan18bmvc,agustsson19cvpr,majumder19cvpr,castrejon17cvpr,hu19nn,maninis18cvpr} and instead define simulations based on more realistic user input.

We present extensive experiments demonstrating that our system can be effectively used by human annotators and brings a considerable reduction in annotation time (Sec.~\ref{sec:experiments}).
In particular we demonstrate on the COCO Panoptic Challenge dataset an overall reduction of over a factor $2\times$ compared to traditional polygon drawing tools ($43\%$ due to our better manual painting interface Mask Paint, and an additional $18\%$ due to the automatic propagation in Magic Paint).
Importantly, these speed improvements come without any loss in annotation quality.
We also show on ADE 20k and Fashionista that our method generalizes to new datasets containing classes not present in its training set.

%% file: fig_teaser.tex
\begin{figure}[t]
\begin{center}
\begin{tabular}{ccc}
  \includegraphics[width=0.32\columnwidth]{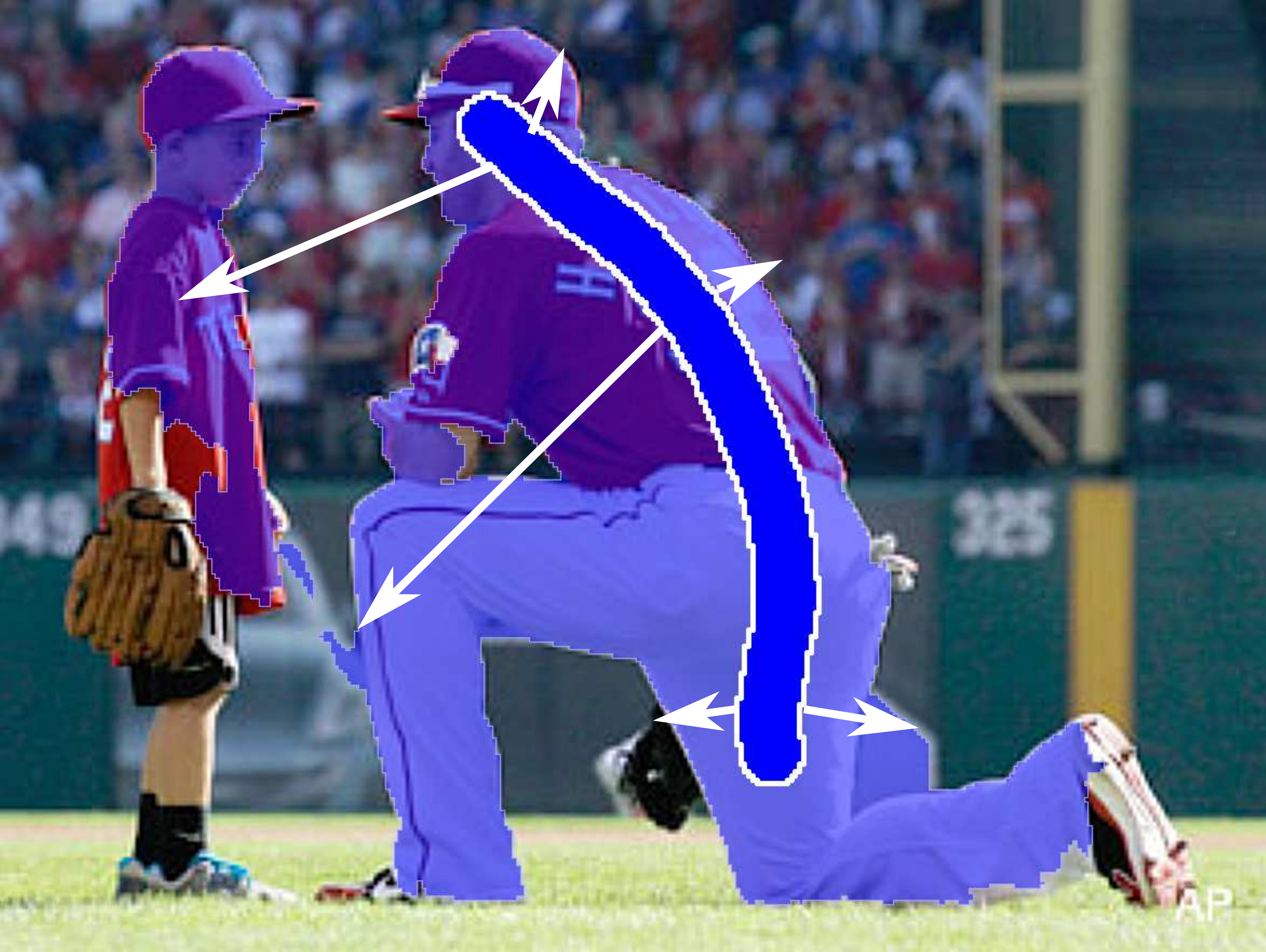}&
  \includegraphics[width=0.32\columnwidth]{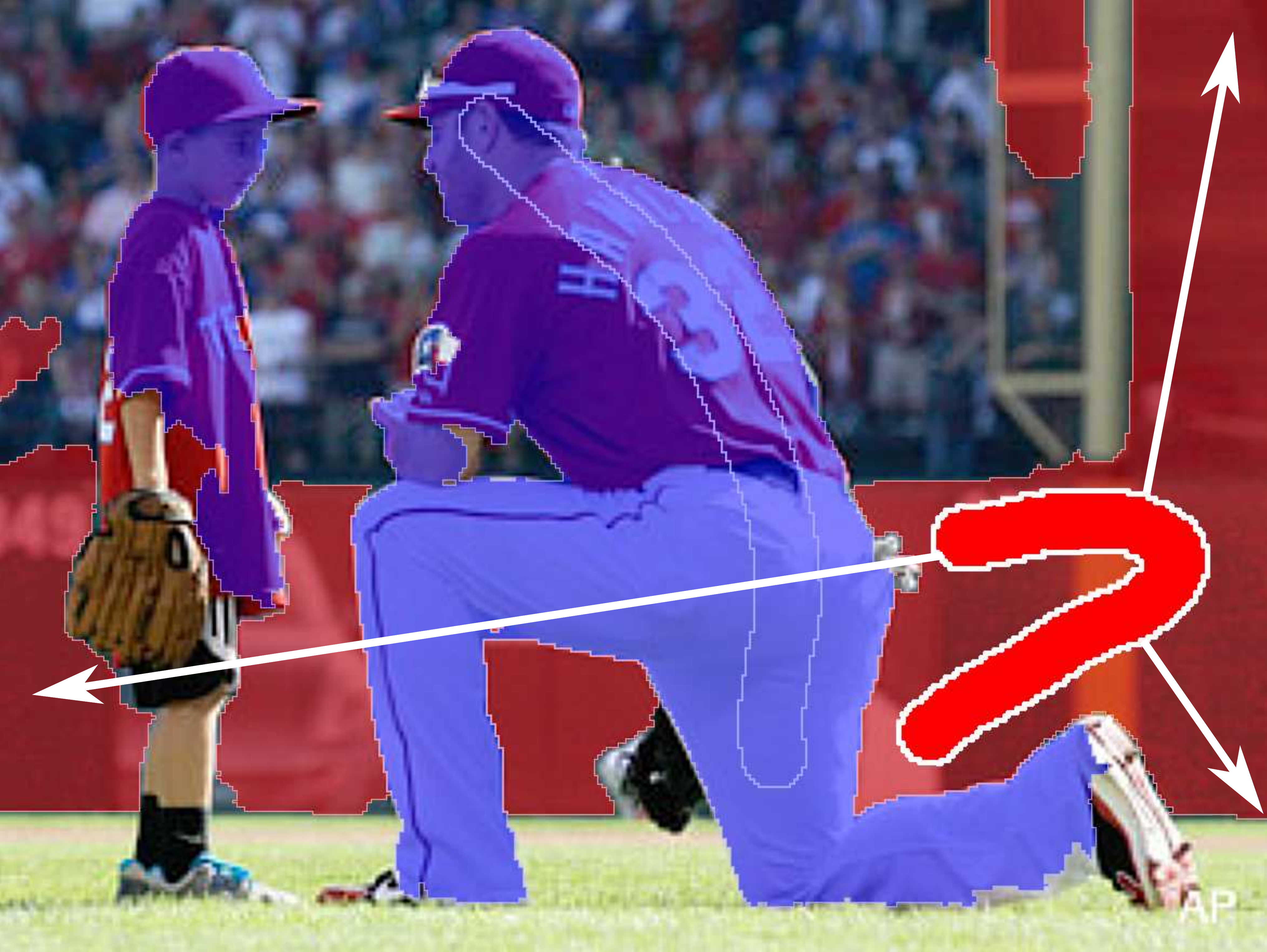} & 
  \includegraphics[width=0.32\columnwidth]{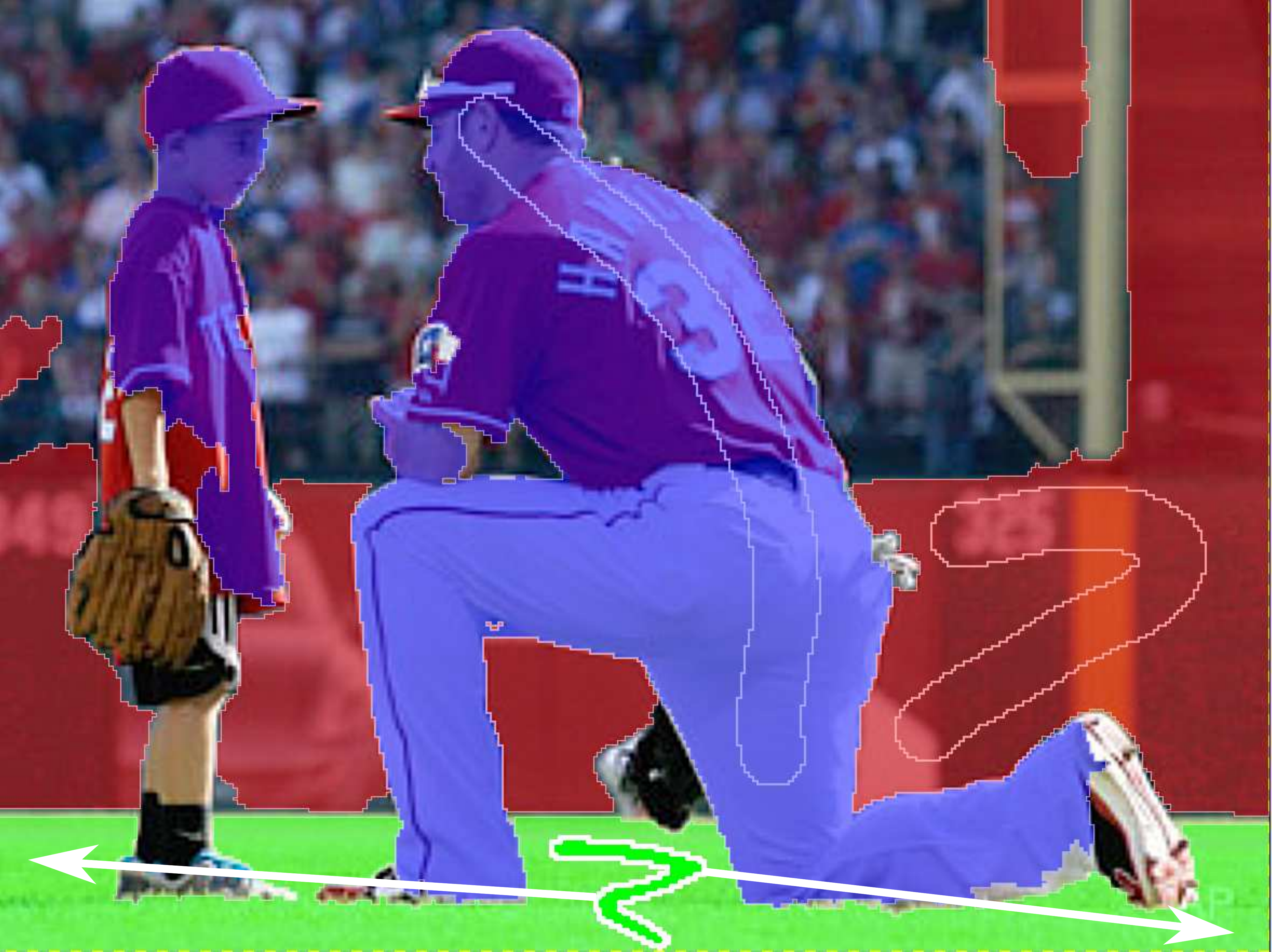}\\
\end{tabular}
\end{center}
\vspace{-0.5cm}
\caption{\small An example of annotation using our Magic Paint approach.
The user first draws a blue stroke on a person, which we automatically propagate to label the majority of that person and the other one (left).
Next, the annotator draws a red stroke, which causes Magic Paint to both correct some minor leaks from the blue region and also propagates correctly to the wall region of the stadium (middle).
Finally a single green stroke by the annotator gets correctly propagated to the entire ground (right).
Note how just three simple strokes already lead to correctly annotating most of this image.}
\label{fig:teaser}
\end{figure}

%% file: related_work.tex
\section{Related work}
\label{sec:related_work}

The goal of an image annotation tool is to provide the underlying media with annotations such as bounding boxes, labels, object masks, etc. Over the course of the years, several interface modalities have been studied: question answering~\cite{deng14chi}, drag and drop UI~\cite{lin14eccv}, touch~\cite{pizenberg2017}, gaze~\cite{papadopoulos14eccv}, and voice~\cite{gygli19cvpr}. The most popular category, which this work belongs to, is that of mouse drawing interfaces.

An annotation tool can simply be manual drawing~\cite{Russell05}, which allows to directly draw the desired annotation on the image (e.g. a bounding-box or a polygon).
While predictable, easy to deploy and intuitive to use, these tools are expensive to employ on larger annotation campaigns. To reduce the annotation effort, the drawing can be machine assisted, like in Polygon-RNN~\cite{acuna18cvpr,castrejon17cvpr}, where a recurrent neural network predicts polygon vertices that the user can interactively correct. Or the focus can be on a more efficient interface, like in the one proposed by~\cite{papadopoulos17iccv} for producing bounding boxes.

A different approach to reduce the annotation cost, in particular for the task of segmentation, has been that of using weak supervision. The weakly labeled data has been of many types: image-level labels~\cite{zhou18cvpr,kolesnikov16eccv}, point clicks~\cite{bearman16eccv,chen18aaai,majumder19cvpr}, object bounding boxes~\cite{khoreva17cvpr,maninis18cvpr}, scribbles~\cite{lin16cvpr,bearman16eccv} or a combination of these~\cite{xu15cvpr}. These weakly supervised methods provide the basis for many interactive annotation tools, where the human collaborates with the machine, usually to verify or correct its output.
Classic interactive approaches for single object segmentation have been proposed in the context of energy minimization over graphs modeling the pixel distribution~\cite{bai09ijcv,batra11ijcv,boykov01iccv,rother04siggraph,criminisi10tog,cheng15cgf,gulshan10cvpr,nagaraja15iccv}.
In the last few years, Fully Convolutional neural networks (FCNs~\cite{chen18pami,long15cvpr}) have been adapted by many approaches for single object segmentation, using point clicks as a form of interaction~\cite{benenson19cvpr,liew17iccv,li18cvpr,mahadevan18bmvc,xu16cvpr,hu19nn,le18eccv,maninis18cvpr,majumder19cvpr,sofiiuk2020cvpr,zheng2020cvpr}.
In~\cite{chen18cvpr} the annotator specifies the object of interest in a video by means of a mask or a set of points. An FCN is used to produce a pixel-wise embedding space, where the non-annotated pixels are classified following a nearest neighbor approach. In~\cite{le18eccv} an FCN takes as input the image and user boundary clicks to predict a boundary map. The final object boundaries are extracted using a minimal path solver~\cite{cohen96cvpr}.

Most of the interactive segmentation works discussed so far focus on segmenting single objects. Recently, a few works have addressed the full image annotation problem~\cite{andriluka18acmmm,agustsson19cvpr}. The approach presented in~\cite{agustsson19cvpr} starts with the annotator specify the four extreme points for each object~\cite{papadopoulos17iccv}. Successive interactions are carried out by correcting the machine prediction with scribbles. In the work of~\cite{andriluka18acmmm} instead, the annotator produces a full image segmentation by composing segments out of a predefined pool generated by the machine.

Part of the technical implementation of our method (Sec. \ref{subsec:embmodel}) is related to the video object segmentation approach~\cite{voigtlaender2019cvpr}. That method matches pixel-wise embeddings to the previous frame predictions and to the ground truth of the first frame to produce per-object distance maps. These are then fed to a segmentation head.
Our works differ in several ways:
we operate in an interactive setting,
we tackle full-image segmentation including background regions rather than not only objects,
and we propagate across different regions in same image showing different instances of the same class, rather than across frames showing the same identical object in a video.
Another technically related work~\cite{fathi2017arxiv} computes instance segmentations by sampling predicted object seed points and using an embedding space to expand the seeds into object masks. We train the embedding space in a similar way, but in our case (i) the seeded points are replaced with actual annotations and (ii) we address the full-image segmentation problem, rather than instance segmentation.

%% file: mask_painting.tex
\section{Annotation Interface}
\label{sec:annotation_interface}

Our system has two main components. A manual drawing interface and an automatic assistant. The manual interface can be used independently.
The assistant component uses the manual annotations and the underlying image to propagate pixel labels across the image. We call the annotation with manual interface alone `Mask Paint', while when there is also the assistant component we call it `Magic Paint'.
Below we describe the two components in detail.

\input{fig_ui}

\vspace*{-0.1cm}
\subsection{Mask Paint: The manual annotation interface}
\label{subsec:manualtools}

Designing an annotation interface requires choosing the right trade-off between simplicity and generality. An interface that allows only to use simple drawing primitives, such as points~\cite{bearman16eccv} or polygons~\cite{russell08ijcv}, has the advantage of being intuitive to use. However, these simple primitives might not be suited for annotating certain types of objects. For example labeling with polygons is problematic for object classes with holes or thin components. A general interface instead should cope with a large variety of object classes. The added complexity needs to be compensated with proper rater training (see Sec.~\ref{subsec:humanexp}), to minimize the chance of the annotator misusing the interface (and ultimately being slow).

We decided to design a general and versatile interface. The user first select one \textit{active label}, which corresponds to a color and a class name (Fig.~\ref{fig:ui}). We allow for three types of drawing primitives (free-form drawing, line drawing, and flood filling), and offer several convenient features (dynamic cursor size, freeze foreground, undo/redo).

\myparagraph{Free-form drawing:} assigns the active label to the pixel under a stroke drawn while holding the left mouse button. It can be used to quickly draw strokes inside an object or to trace curved object boundaries.

\myparagraph{Line drawing:} assigns the active label to the pixels under a straight polyline. It can be used to delineate object contours, as in polygon labeling tools such as \cite{russell08ijcv}. It can also be used to quickly draw narrow elongated object segments.

\myparagraph{Flood filling:} assigns the active label to a bounded region of the annotation map which has an existing uniform label or is a background region (i.e. currently unassigned to any label).

\myparagraph{Dynamic cursor size:} The stroke size for both free-form and line drawing can be dynamically adjusted by scrolling the mouse wheel. They can also be used to erase previously labeled pixels by using the right mouse button.

\myparagraph{Freeze foreground:}
This special drawing mode modifies the behavior of the free-form and line primitives.
It assigns the active label to the region defined by the stroke if the region has no label yet (\textit{i.e.}, background). Otherwise it \textit{confirms} the current label (this will be relevant for Magic Paint, see Sec.~\ref{subsec:autotools}).
This mode allows drawing the boundary separating two objects only once: if one object has been already annotated, we can safely paint over its boundary with a different label and color only the background side of the boundary. This is similar to the superpixel-based labeling of stuff classes after freezing foreground classes in \cite{caesar18cvpr}, except our annotations are at the pixel level.

\myparagraph{Undo/redo:}
During annotation we maintain a record of the primitives drawn by the user and use it to implement undo/redo functionality. This record is also used for replaying the strokes in simulation (see Sec.~\ref{subsec:ablationexp} and Sec.~\ref{subsec:otherdataexp}) and to evaluate timing and quality (see Sec.~\ref{sec:experiments}).

We demonstrate in Sec.~\ref{subsec:humanexp} that in combination these features make our interface significantly more efficient for full-image annotation than polygon drawing.

\input{fig_tools}

\vspace*{-0.1cm}
\subsection{Magic Paint: Annotation with automatic label propagation}
\label{subsec:autotools}

The second component of our interface automatically propagates pixel labels after each human annotation action (Fig.~\ref{fig:teaser},~Fig.~\ref{fig:tools},~Fig.~\ref{fig:example_results}).
To describe the interaction of this component with the human, we need to define two types of segmentations that coexist during the annotation process. The first is the \textit{reference} segmentation, which is made by the human (or confirmed by the human, see below).
The second is the \textit{predicted} segmentation made by the assistant.
The assistant can never modify the reference segmentation. It can only propose a label for unlabeled (\textit{i.e.}, background) pixels or change pixel labels within a predicted region.
Furthermore, the assistant bases its predictions only on the reference segmentation, as their labels are considered correct.

A predicted label can be \emph{promoted} to reference in two ways.
The first is for the human to flood fill a predicted region to change its label or to confirm its predicted label (Fig.~\ref{fig:tools}, top row, middle).
The second is through confirming a region underlying a stroke of the freeze-foreground. This is particularly beneficial when drawing over boundaries between two regions, where one is background and the other is a correctly predicted segmentation (Fig.~\ref{fig:tools}, top row, right). The background gets assigned the active label, while the predicted pixels touched by the freeze-foreground stroke get confirmed and promoted to reference.
The newly promoted pixels will be used to generate reference examples to base further label propagation on.

The algorithm used by the assistant to propagate pixel labels will be presented in the next Section (Sec.~\ref{sec:model}).

%% file: fig_ui.tex
\begin{figure}[t]
\begin{center}
  \includegraphics[width=0.47\columnwidth]{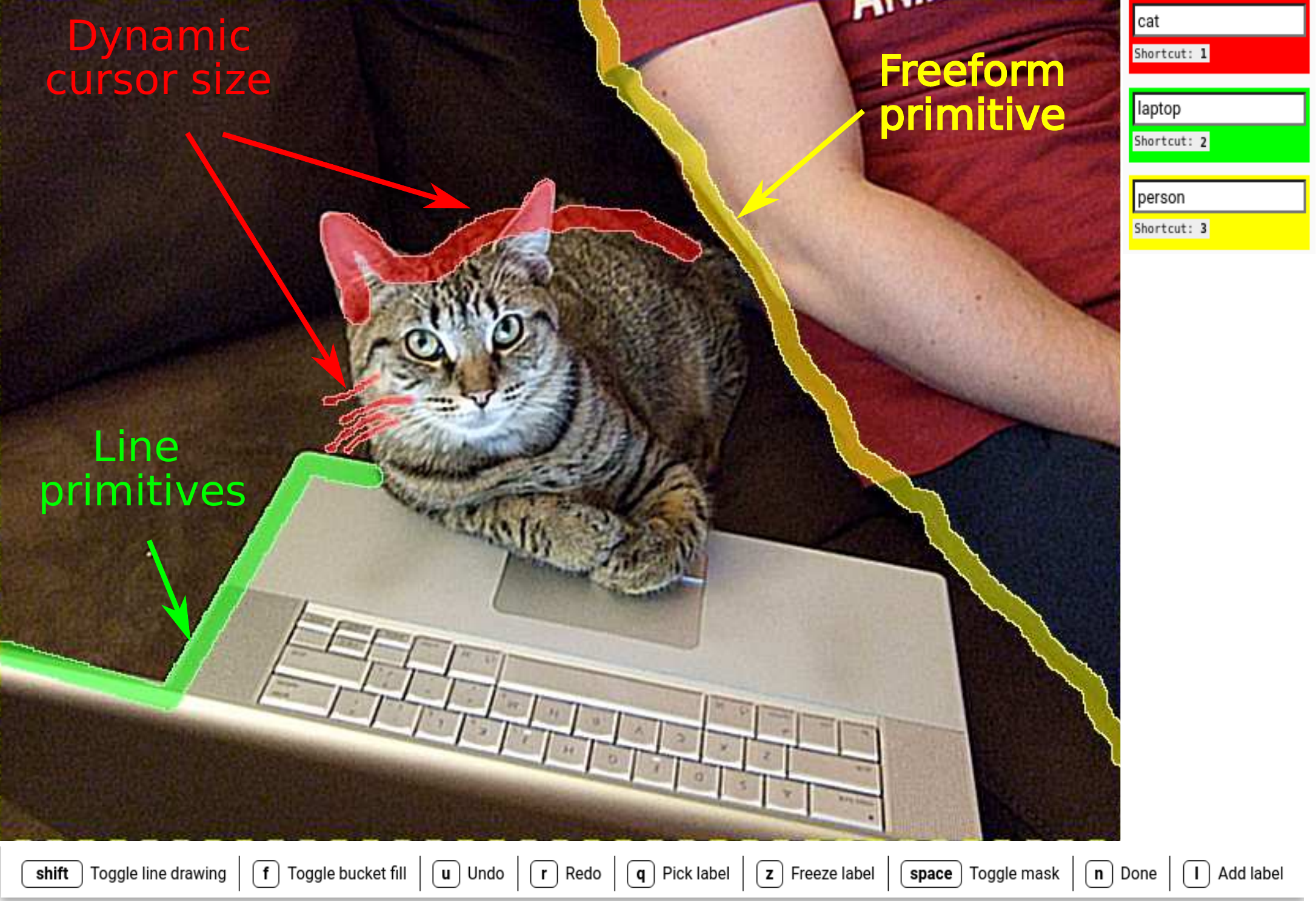}
  \includegraphics[width=0.47\columnwidth]{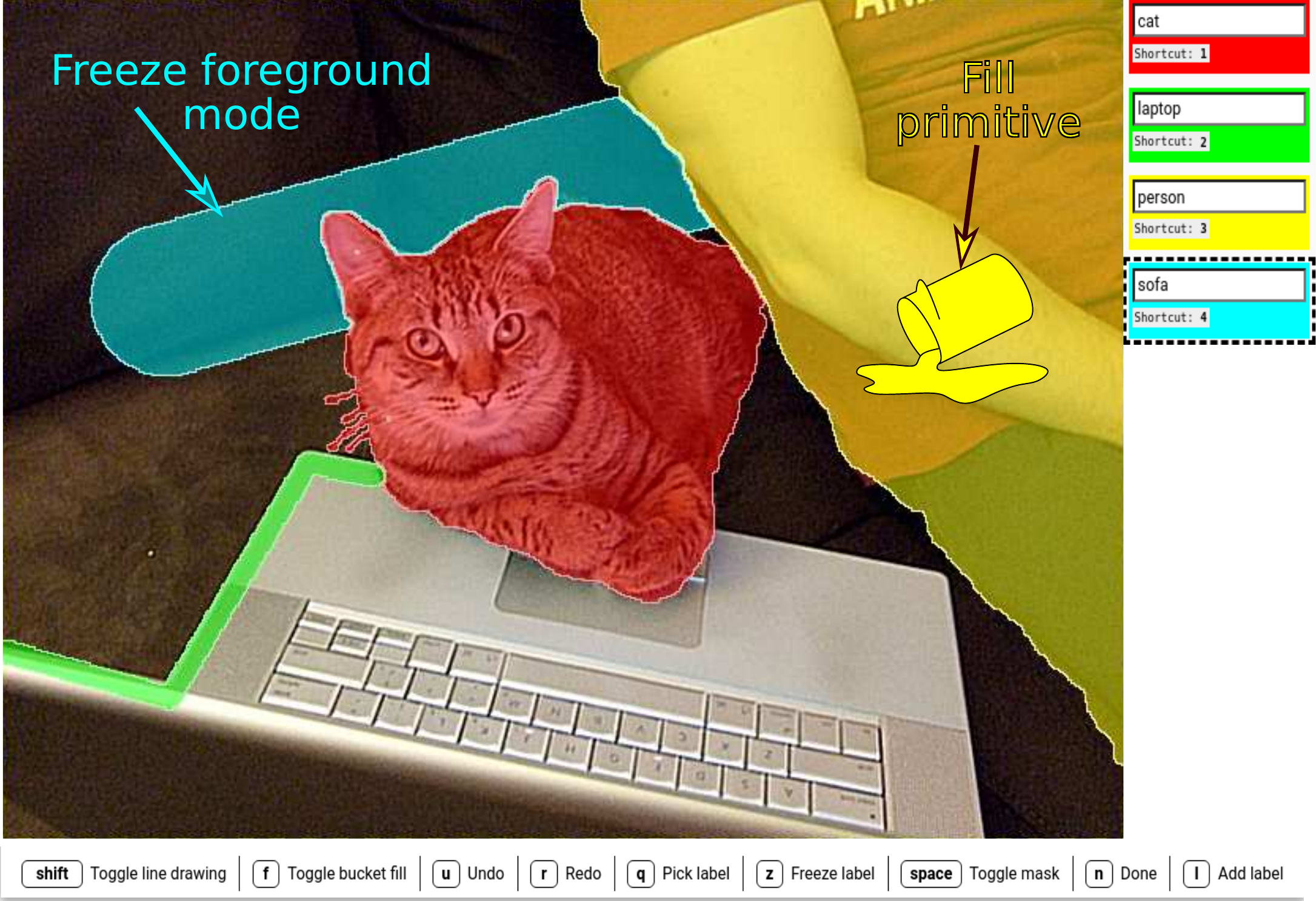}
\end{center}
\vspace{-0.5cm} 
\caption{The Mask Paint UI.
The left figure shows the line (green) and freeform primitives (yellow, red). The cursor size can be adjusted according to the desired degree of detail (red).
In the right figure, the person region has been filled with the flood fill primitive (yellow), while the sofa is being annotated in the freeze foreground mode (light blue), to avoid drawing the cat-sofa boundary twice.}
\label{fig:ui}
\end{figure}

%% file: fig_tools.tex
\begin{figure}[t]
\begin{center}
\begin{tabular}{cccccc}
  \multicolumn{2}{c}{
    \includegraphics[width=0.32\columnwidth]{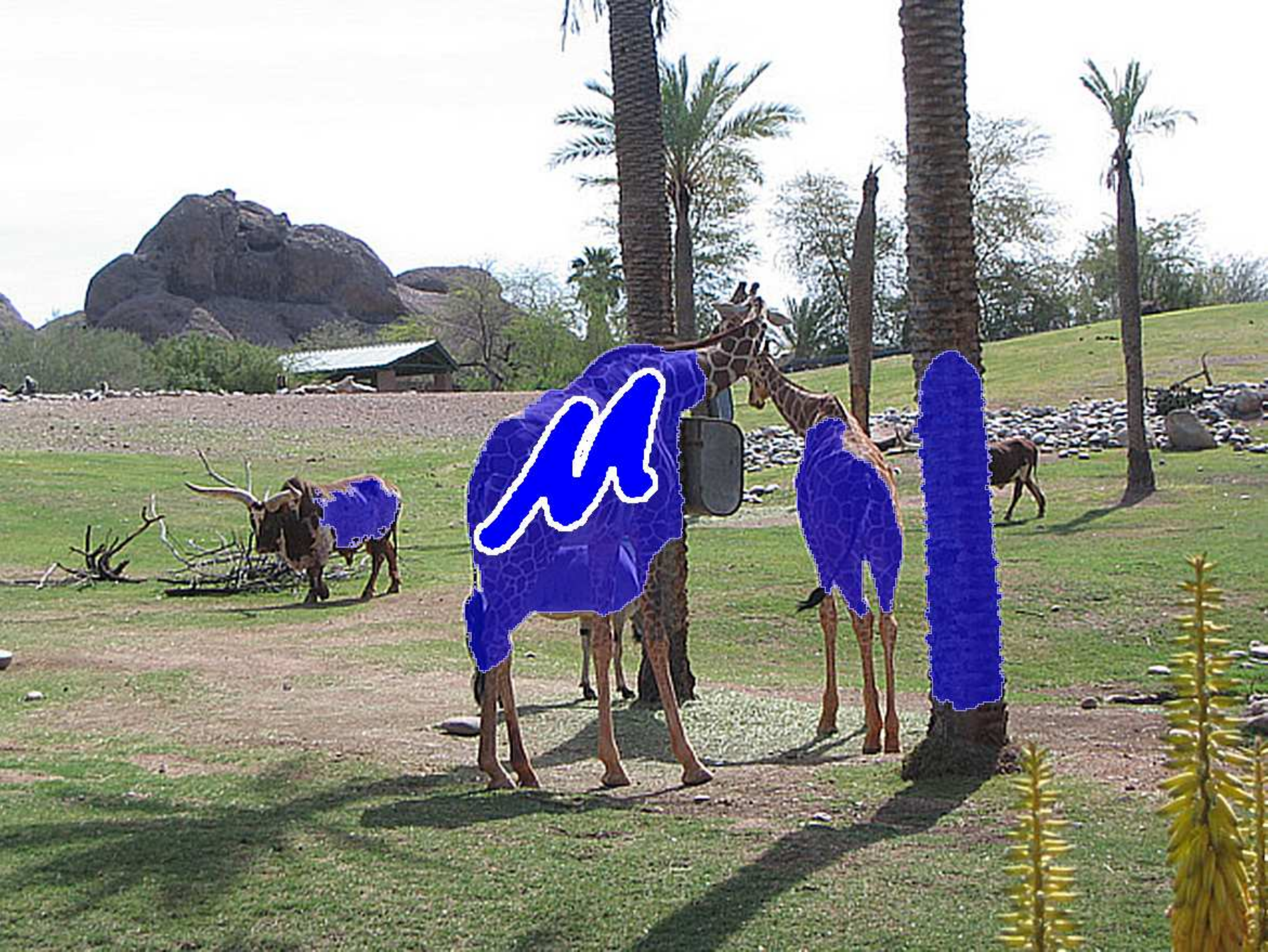}
   }&
  \multicolumn{2}{c}{
    \includegraphics[width=0.32\columnwidth]{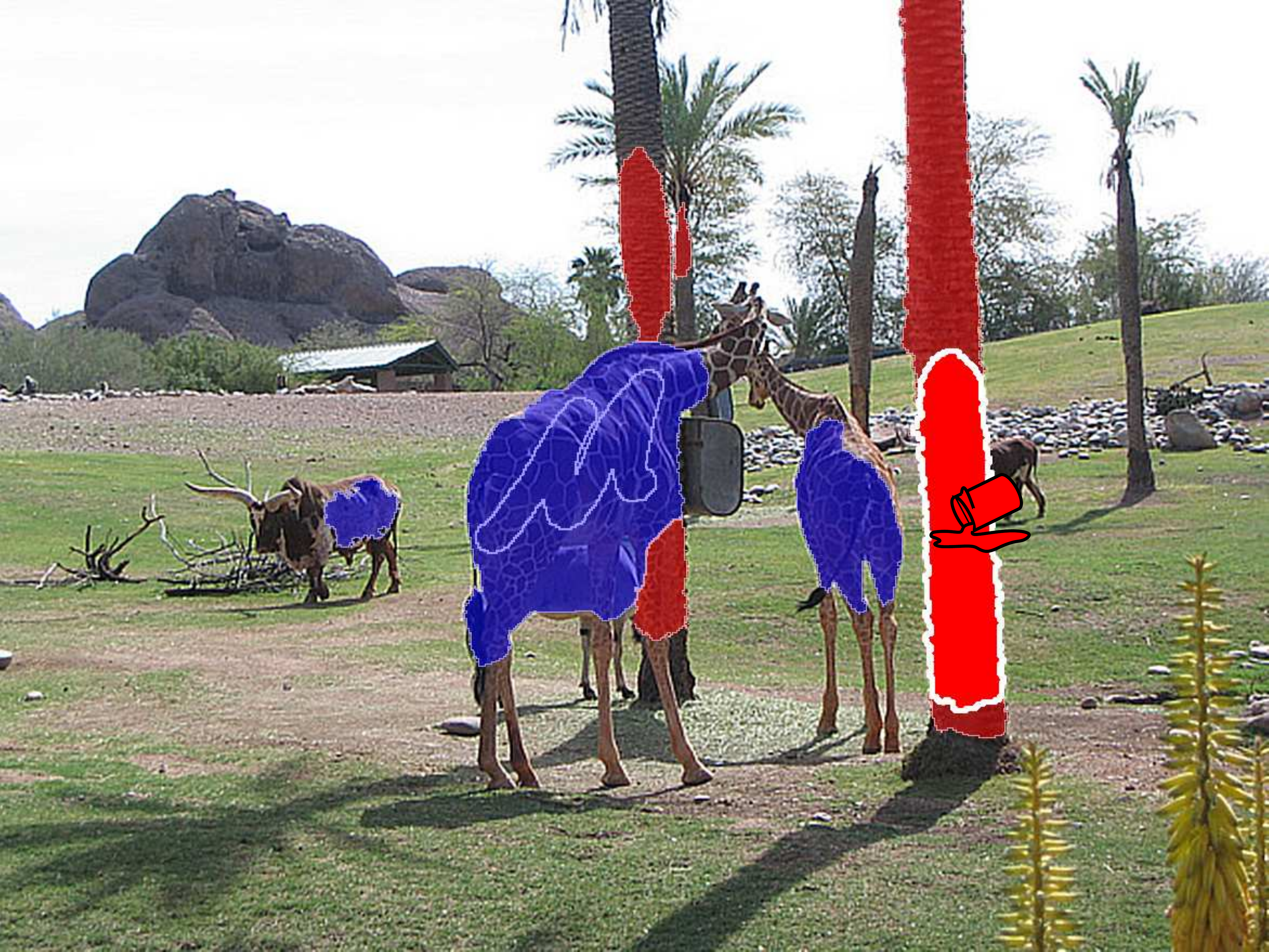}
  }&
  \multicolumn{2}{c}{
    \includegraphics[width=0.32\columnwidth]{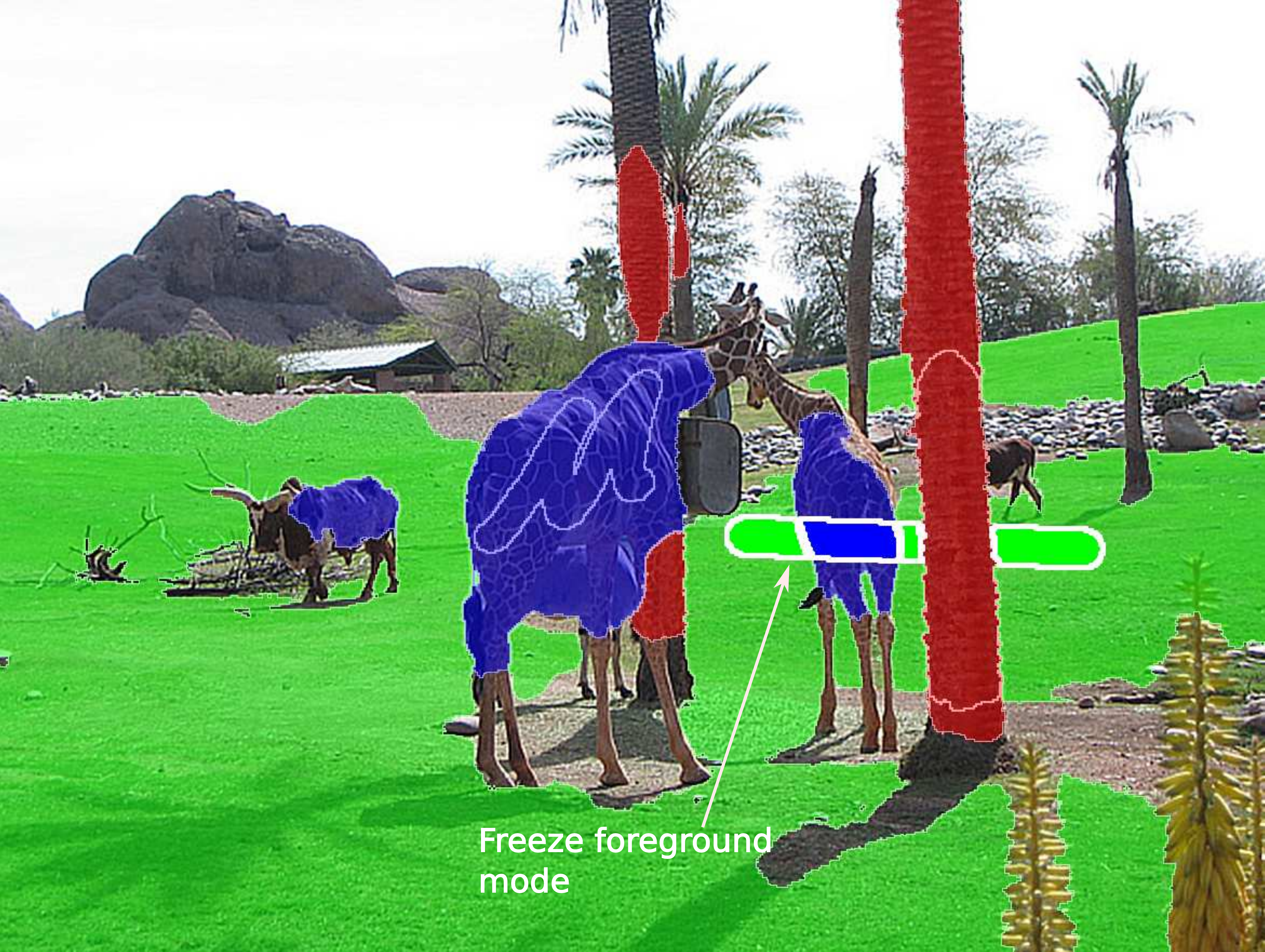}
  }\\
  \multicolumn{2}{c}{
    \includegraphics[width=0.32\columnwidth]{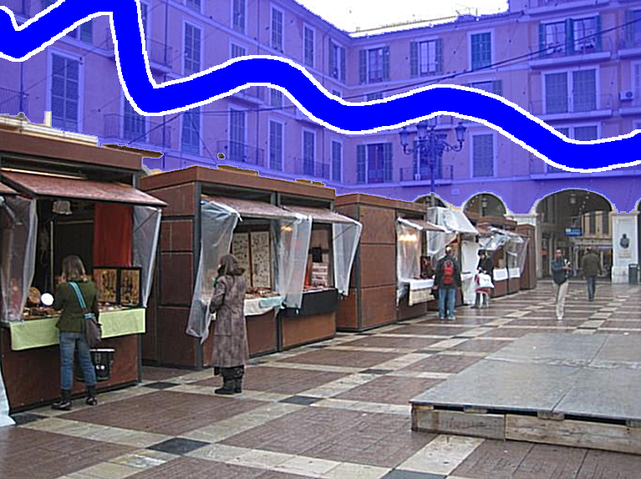}
  }&
  \multicolumn{2}{c}{
    \includegraphics[width=0.32\columnwidth]{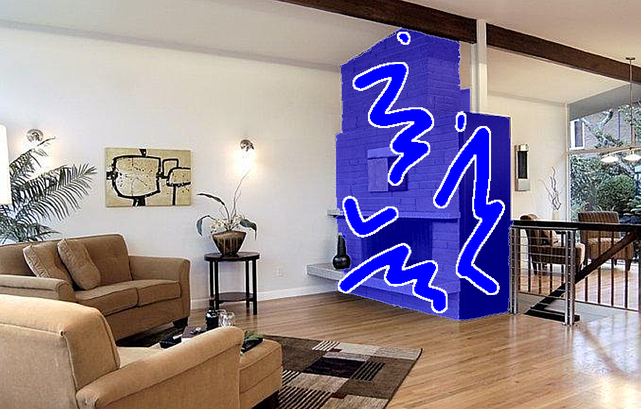}
  }&
  \includegraphics[width=0.16\columnwidth]{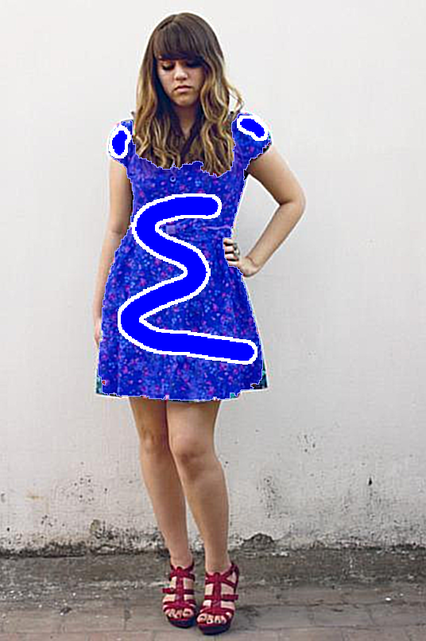} &
  \includegraphics[width=0.16\columnwidth]{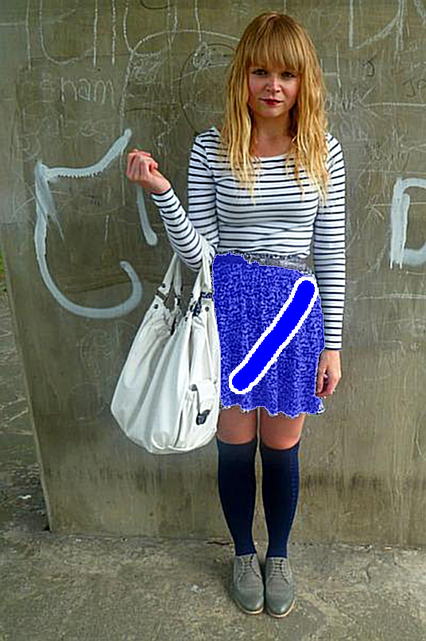} \\
\end{tabular}
\end{center}
\vspace*{-0.5cm}
\caption{Magic Paint usage. \textbf{Top Row}.
Left: Magic Paint propagates a stroke to label more of the giraffes, but it also leaks onto the tree.
Middle: the annotator corrects the leak by filling it with a different label (red). Magic Paint then automatically annotates more tree surface correctly.
Right: a single stroke, in freeze foreground mode, is propagated to the entire ground. Note that the background stroke, while coloring green the unlabeled pixels, also \emph{confirms} the blue pixels of the giraffe that it touches.
\textbf{Bottom Row}. More examples of propagations. The first two images come from the ADE-20K dataset~\cite{zhou17cvpr} and the stroke propagates through the "building" and "fireplace" classes. The last two images come from the Fashionista dataset~\cite{yamaguchi12cvpr} and the stroke annotates the classes "dress" and "skirt". The last three classes are not present in the COCO Panoptic dataset on which Magic Paint was trained (fireplace, dress, skirt).
}
\label{fig:tools}
\end{figure}

%% file: model.tex
\section{Similarity Based Propagation Algorithm}
\label{sec:model}

\input{fig_architecture}

Let us denote the input image as $I$ and a set of partial annotations provided at a certain point during the annotation process as $\{A_n\}$, where $n \in \{1, \ldots, N\}$ and $N$ is the total number of
classes in the partial annotation. Each $A_n$ is represented as binary map containing the annotator strokes.
The goal of the annotation assistant is to estimate the complete labeling $L=\{l_k\}$ of the image, where $l_k \in \{0, \ldots, N\}$ is the
label of pixel $k$ and $l_k=0$ indicates that assistant is not confident enough to provide an estimate.

Ideally we would like to propagate information contained in the labeled pixels $A_n$ over the entire image. However we would also like to make predictions locally coherent to reduce clutter and ease the cognitive burden on the annotator.
To strike a trade-off between global propagation and local consistency we propose a system that
operates in two steps. In the first step we globally compare labeled and unlabeled pixels across the
entire image based on their appearance similarity.  This already provides estimates for the labels
of unlabeled pixels, and does not depend on proximity to the labeled area, hence achieving maximum
propagation effect.
In the second step we use the results of global similarity computation 
as input into an inference model that estimates the class labels of all pixels jointly.
The inference model imposes a spatially coherent labeling and it helps adhering to image boundaries.
In the following we proceed by describing our pixel similarity model and the inference model considered in our experiments. 

\vspace*{-0.1cm}
\subsection{Pixel similarity model}
\label{subsec:embmodel}

We rely on pixel embeddings as appearance representation for global similarity computation. 
To that end we introduce a convolutional neural network module $E$ which given an image $I$ outputs an embedding vector for every pixel.
We train $E$ following \cite{fathi17arxiv,siyang18cvpr} but repurpose the training setup to obtain embeddings dedicated to work well for measuring within-image appearance similarity.

For a pair of pixels $(i, j)$ the pairwise similarity between their embeddings is defined as
\small
\begin{equation}
  \sigma_{ij} = \frac{2}{1 + \exp(\|\bve_i - \bve_j\|_2^2)}
\end{equation}
\normalsize
where $\bve_k$ is the embedding vector of a pixel $k$. The embedding loss for a pair $(i,j)$ is then given by
\small
\begin{equation}
  \mathcal L_{ij} = \mathbbm{1}_{ij}\log\sigma_{ij} + (1 - \mathbbm{1}_{ij})\log(1- \sigma_{ij}), 
\label{eq:embloss}
\end{equation}
\normalsize
where ${1}_{ij}$ indicates whether both pixels $i$ and $j$ belong to the same class.

\myparagraph{Model details.}
We use a lightweight model to keep it computationally cheap and suitable for interactive
experiments. Moreover, this also prevents overfitting to a specific dataset.
it is composed of $7$ convolutional layers with kernel sizes equal to $3$ and increasingly large number of filters in each layer given by $[32, 64, 128, 256, 256, 256]$.
We compute all convolutions with stride $1$ and use padding to keep this resolution throughout all the layers in our model. As in \cite{chen17arxiv} we rely on dilated
convolutions instead of downsampling to expand the size of the receptive field.

\myparagraph{Training the model.}
Given a dataset of images with semantic segmentation ground-truth we train $E$ by minimizing the loss given by Eq.~\eqref{eq:embloss} averaged over a set of pixel pairs. To generate a batch of training examples we first sample a fixed number of images and pass them through $E$ to obtain a set of dense pixels embeddings. We
then sample a fixed number of pixel pairs from each image in the batch\footnote{In our
  experiments in this paper we train the embedding network with 5 images per batch and $500$
  pixel pairs per image.}.
  We rescale all images to $300\times 400$ pixels beforehand.
  Note that $E$ is simultaneously trained on pairs from diverse set of
classes which encourages it to learn generic features useful to determine ``same
class'' membership.
We use the Adam optimizer~\cite{kingma15iclr} with learning rate $10^{-4}$.

\myparagraph{Distance maps on a test image.}
After training $E$, we can use it on a new test image for which some manually labeled pixels are available.
Following \cite{voigtlaender2019cvpr} we represent the output of the global pixel similarity computation
 between the labeled and unlabeled pixels in the form of distance maps. The distance map $D_n$ for a class $n$ at pixel $k$ is given by
\small
\begin{equation}
  D_n(k) = \min_{j \in \bar{A}_n}\|\bve_k - \bve_j\|_2^2,
\end{equation}
\normalsize
where $\bar{A}_n$ is a set of all manually labeled pixels in the current annotation map $A_n$.
We define a distance map $D_0$ to represent the background class and we set it to a constant value.

\vspace*{-0.1cm}
\subsection{Inference model}
\label{sec:mask-prediction}

Given the distance maps $\{D_n\}$ the simplest way to predict a label $l_k $ for a pixel at location $k$ is 
\small
\begin{equation}
  l_k = \argmin_{n\in\{0, \ldots, N\}}D_n(k),
\label{eq:onenn}
\end{equation}
\normalsize
which corresponds to a nearest neighbor classifier evaluated independently for each pixel.

\myparagraph{DenseCRF Model.}
We consider here a better alternative for inferring pixel labels that can produce spatially coherent
predictions jointly for all pixels.
For this we apply a dense conditional random field model \cite{kraehenbuehl11nips}, which brings two advantages (Fig. ~\ref{fig:architecture}):
(1) the CRF will smooth-out spurious label switches in the middle of homogeneous regions that a nearest-neighbor assignment based purely on individual embedding distances would sometimes produce;
(2) the CRF will produce more accurate object boundaries, by reacting to low-level image gradients between neighboring pixels; in contrast the embedding model might over-smooth the boundaries as it integrates information across them.

We define a CRF graph, where each node is a pixel and there is an edge between any two pixels. The label set is defined by the set of labels for which there is at least one annotation in the image, plus a background label. The energy $E(L)$ over the label assignments is defined as
\small
\begin{equation}
E(L) = \sum_k \psi_u(l_k) + \sum_{k<j}\psi_p(l_k, l_j)
\end{equation}
\normalsize
where $\psi_u$ and $\psi_p$ are the unary and pairwise potentials. The unary potentials assign the value of the distance map for label $l_k$ at pixel $k$, {\em i.e.}, $\psi_u(l_k) = D_{l_k }(k)$.
The pairwise potentials use are a sum of smoothness and appearance similarity kernels, as in \cite{kraehenbuehl11nips}:
\small
\begin{equation}
\psi_p(l_k, l_j) = \exp(-\frac{\|\mathbf{p}_k - \mathbf{p}_j\|_2^2 }{\theta_{\gamma}} ) + \alpha\exp(-\frac{\|\mathbf{p}_k - \mathbf{p}_j\|_2^2 }{\theta_{\alpha}} -\frac{\|\mathbf{I}_k - \mathbf{I}_j\|_2^2 }{\theta_{\beta}} ),
\end{equation}
\normalsize
\noindent where $\mathbf{p}_i$ is the position of pixel $i$ and $\mathbf{I}_i$ is its RGB color
vector. We use grid search to identify good values for $\alpha$, $\theta_\gamma$, and $\theta_\alpha$ on a validation set.

%% file: fig_architecture.tex
\begin{figure}[t]
\vspace{-0.3cm}
\begin{center}
  \includegraphics[width=0.99\linewidth]{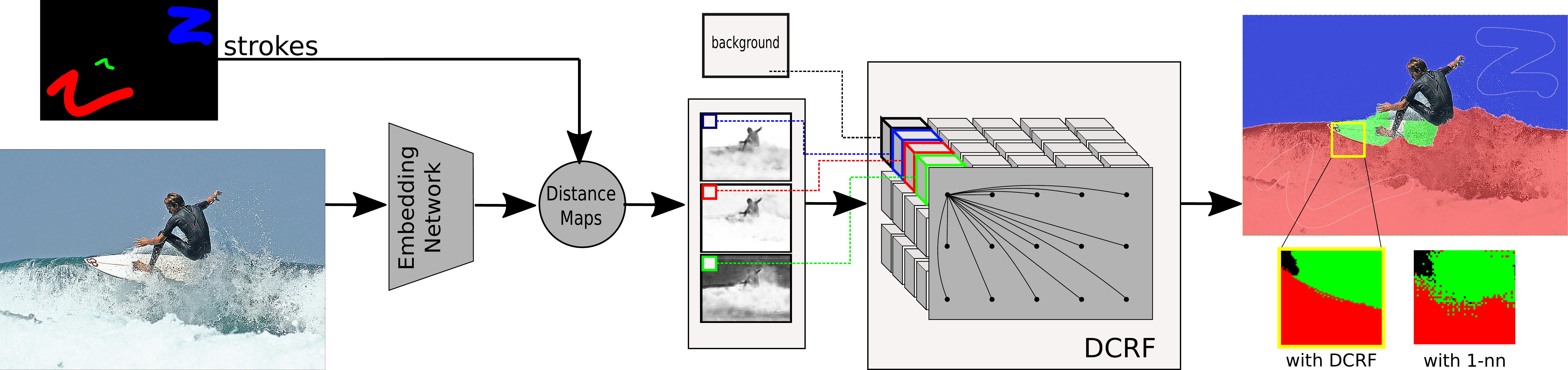}
\end{center}
\vspace{-0.5cm}
   \caption{The architecture of the label propagation assistant. The annotator provides a stroke for the class "sea" (red), one for "surfboard" (green) and one for "sky" (blue). The pixel embeddings are combined with the annotator's strokes to generate one distance map for each class. The distance maps values are then used as unary terms within a DCRF to produce the final segmentation.
   Note the effect of the DCRF in the magnified region of the output on the right, when compared to 1-nearest-neighbor classification in embedding space (1-nn)}
\label{fig:architecture}
\end{figure}

%% file: experiments.tex
\section{Experiments}
\label{sec:experiments}

We evaluate our Mask Paint and Magic Paint annotation approaches on several
established computer vision datasets. We include both simulations and experiments
with real human annotators. We evaluate annotation accuracy using average
intersection over union (IoU) as is common in the literature
\cite{chen18pami,benenson19cvpr,agustsson19cvpr,shelhamer16pami,zhou17cvpr,zheng2020cvpr}.
When comparing two annotation approaches we consider both annotation accuracy and the time required to reach it.

\input{fig_example_results}

\myparagraph{Simulation environment for Magic Paint.}
Creating a full simulation environment for Magic Paint is challenging because the underlying Mask Paint UI allows for a large variety of possible annotations, which depend on the content of a particular image and the annotated classes. For example, the annotator can decide whether to annotate the contour of an object first and then flood-fill its interior, or use thicker strokes to annotate it in one go. Such behavior is difficult to simulate based on full-image segmentation ground-truth annotations alone. Moreover in a fully automatic simulation it is not clear how much time each annotation action should take.
We therefore propose to simulate Magic Paint based on recordings of real human annotators using Mask Paint, keeping track of both annotation actions and their timing.
At simulation time we replay the annotation recording to reconstruct the annotation process.
Each time we replay a recorded stroke we feed the resulting partial annotation to the assistant, which then estimates the labels of all unlabeled pixels (Sec.~\ref{sec:model}).
Prior to executing the next recorded annotation action, we check whether its effect was already covered by the assistant.
More precisely, we declare a new action as redundant and skip it if it does not improve the average IoU of the image segmentation w.r.t to the ground-truth. This simulates the fact that the user, recognizing that the area of the stroke was already correctly estimated by the assistant, would not do it at all.
Since during the annotation process the assistant might change the labeling of previously estimated
pixels we perform multiple passes through the recording and at each pass execute the remaining non-redundant actions.
Note that our simulator is only applicable to recordings in which
actions are independent of each other. Such assumption is violated by actions that do
``freeze-foregrund'' and ``flood-fill''.
To evaluate our assistant in simulation nonetheless we collect human annotation recordings without these actions in addition to those made with the
full Mask Paint.

Our methodology can measure improvements in terms of annotation time, rather than merely in the number of annotator actions (e.g. the number of clicks~\cite{mahadevan18bmvc,andriluka18acmmm,agustsson19cvpr}), which is more relevant since some actions can take longer than others. Another advantage is that both the timing and the appearance of the strokes are real, which would be difficult to achieve in a fully automatic simulation.

\vspace*{-0.1cm}
\myparagraph{Datasets.}
We train the pixel embedding model used in all our experiments on a subset of $30000$ images from
the training set of the COCO Panoptic dataset \cite{cocopanoptic,caesar18cvpr} (``COCO-train-30k'').
We keep this model fixed in all experiments, also when evaluating on other datasets.

We start by validating the performance of the components of our Magic Paint system on COCO Panoptic (Sec.~\ref{subsec:ablationexp}).
For this, human annotators label a random subset of $100$ images from the validation set with Mask Paint (``COCO-val-100''), and then we use their annotation recordings for simulating Magic Paint.
Next, we perform experiments with human annotators using Magic Paint on this dataset in Sec.~\ref{subsec:humanexp}, and compare to them using Mask Paint and Polygon drawing.
Finally, we also report results on two additional datasets with diverse set of classes and images: ADE-20k \cite{zhou17cvpr} and Fashionista~\cite{yamaguchi12cvpr}. We again manually annotate a random subset of $100$ images from each dataset with Mask Paint and report simulations for Magic Paint (``ADE-val-100'' and ``Fashionista-val-100'').
Several examples of our annotation tool dealing with these datasets are shown in Fig.~\ref{fig:example_results}.
By keeping the embedding model fixed, we demonstrate that our method generalizes to new datasets showing classes not seen during training.

We will make our annotation recordings publicly available to enable reproducible research and comparison to our work.

\myparagraph{Human annotators.}
We work with a team of professional annotators who routinely perform annotation tasks as their daily job. We did not have direct access to the annotation team, and communicated with the annotators solely through instructional materials and video demonstrations.
We required each annotator to pass a qualification task which verified they are able to deliver sufficiently accurate annotations within predefined time limits.
We administer a dedicated training procedure for each interface we evaluate in our experiments, i.e. Mask Paint, Magic Paint, and Polygon drawing (see below). Hence, the timings we report are for annotators that are already well trained in each interface.
To avoid the complexity of dealing with ambiguities in annotation we presented a reference showing ground-truth labeling. This allows the experiment to focus on the effectiveness on the annotation tool and not on orthogonal factors such as the ability of annotators to identify various visual classes in the image.

\vspace*{-0.1cm}
\subsection{Comparison of Magic Paint variants on COCO Panoptic}
\label{subsec:ablationexp}

We start by evaluating the components of Magic Paint in simulation on the
COCO-val-100 dataset.
As detailed above, these simulations are based on records of real human annotators operating a basic Mask Paint interface with just line and free-form tools enabled (``basic'').
We compare our full Magic Paint model to a variant that does not use the inference model and just labels pixels with 1-nearest-neighbor classification in the embedding space (``1-nn'').

We consider two types of pixel embeddings:
(1) the learned embedding model of Sec.~\ref{subsec:embmodel} 
and (2) a baseline embedding based on color histograms in hue and saturation space ($10$ dimensions each).
We use SLIC superpixels to define support for color histograms to encourage the output
to respect image edges.

\input{fig_sim_and_human_coco}

The results of the mode comparison are shown in Fig.~\ref{fig:eval_coco}-left. We observe that
``Magic Paint (1-nn)'' with the learned pixel embedding model improves over using color
histograms. However both ``1-nn'' models considerably underperform compared to the full Magic Paint
with the inference model (``DenseCRF''). Visually inspection reveals that the
``1-nn'' models often produce spurious predictions which need to be corrected by annotator at the cost of additional annotation time. Instead, the CRF produces smoother labelings in homogeneous areas, and yields predictions that are better aligned with image boundaries.

As a final experiment we consider the influence of the ordering in which
actions are process by the simulator. We consider the original annotator ordering (``sim-seq.'') and compare it to randomly sampling actions form the recording ( ``sim-rnd.'').
We observe that ``sim-rnd.'' leads to substantially better results. The reason is that annotators using Mask Paint often proceed by first labeling the outlines of objects, and fill in the interiors later on. This turns out to be suboptimal for Magic Paint. Moreover in ``sim-rnd.'' subsequent actions are providing more diverse input for the annotation assistant in Magic Paint to learn the appearance of the objects and backgrounds in the image, whereas in ``sim-seq.'' subsequent inputs are often made on nearby areas with similar appearance.

\vspace*{-0.1cm}
\subsection{Magic Paint with real human annotators on COCO Panoptic}
\label{subsec:humanexp}

We evaluate here four version of our system, always using human annotators:
(1) the basic Mask Paint interface with only lines and free-form strokes  (``Mask Paint (basic)''),
(2) a version that in addition has a freeze-foreground feature,
(3) a version with also flood-fill (``Mask Paint (full)''),
(4) Magic Paint on top of all of the Mask Paint features.
As a baseline we employ a version of Mask Paint which we modify to operate as polygon labeling tool closely resembling LabelMe~\cite{russell08ijcv} and the annotation tools used for labeling of ADE20k \cite{zhou17cvpr} and Cityscapes~\cite{cordts16cvpr} datasets. We refer to this as ``Polygons''.

Fig.~\ref{fig:eval_coco}-right shows the results.
All versions of Mask Paint significantly improve over Polygons in terms of annotation
time without incurring any loss in accuracy (even improving it somewhat).
This is likely because Mask Paint naturally facilitates sharing of boundaries between classes and provides tools for labeling a variety of object shapes, whereas Polygons lack such flexibility.
We also observe that adding both freeze-foreground and flood-fill further improve the efficiency of Mask Paint.
When considering the end of the annotation process (top point of each curve), the full Mask Paint is 43\% faster than traditional polygon drawing (341s vs 598s).
Going beyond, Magic Paint further improves the total annotation time over the full Mask Paint by about 18\% (280s), again without any loss in quality.
When considering points on the curves corresponding to lower quality targets or fixed annotation budgets, the improvements are even larger, e.g.
(1) Magic Paint reaches 0.6 average IoU 36\% faster than the full Mask Paint, %
and (2) after 200 seconds of annotation per image, Magic Paint delivers +0.2 higher average IoU. %
These results demonstrate that our automatic label propagation indeed helps and that real human annotators are able to take advantage of it.
Note how the timing achieved by human annotators using Magic Paint is very close to the corresponding simulation result from (``Magic Paint, sim-seq. (DenseCRF)'' in Fig.~\ref{fig:eval_coco}-left, 274s vs. 280s).
Overall, we highlight that Magic Paint reduces annotation time by more than a factor of $2\times$ compared to Polygons (280s vs. 598s).

\vspace*{-0.1cm}
\subsection{Annotation of new visual classes on ADE-20k and Fashionista}
\label{subsec:otherdataexp}
\input{fig_transfer}

We now evaluate how Magic Paint performs on datasets with new visual classes not seen during training (simulations on top of recordings of human annotators using the basic Mask Paint).
For this we use the ADE-val-100~\cite{zhou17cvpr} and Fashionista-val-100~\cite{yamaguchi12cvpr} datasets. 
The results in Fig.~\ref{fig:eval_transfer} show that on both datasets 
Magic Paint leads to finishing the annotation substantially faster than with Mask Paint, while reaching the same
quality level.
For example on ADE-20k Magic Paint finishes annotation after 323 seconds, whereas Mask Paint requires 485 seconds.
As our Magic Paint simulation is based on recordings of Mask Paint made without annotators observing label propagation, they represent a conservative estimate of the speed of human annotators using Magic Paint.
For example randomly reordering annotation actions leads to further efficiency gains (compare the ``sim-rnd'' and ``sim-seq.'' curves).
Overall these results show that the learned pixel embeddings and CRF inference model in our system are able to generalize to classes not seen during training. 

%% file: fig_example_results.tex
\begin{figure}[t]
\begin{center}
\begin{tabular}{ccccc}
  \includegraphics[height=2.73cm]{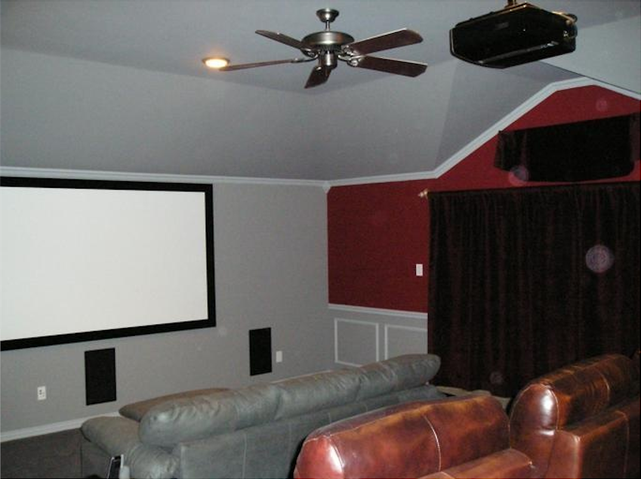}&
  \includegraphics[height=2.73cm]{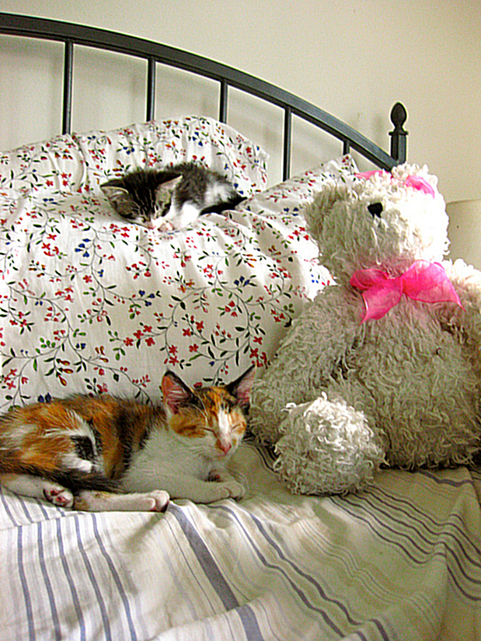}&
  \includegraphics[height=2.73cm]{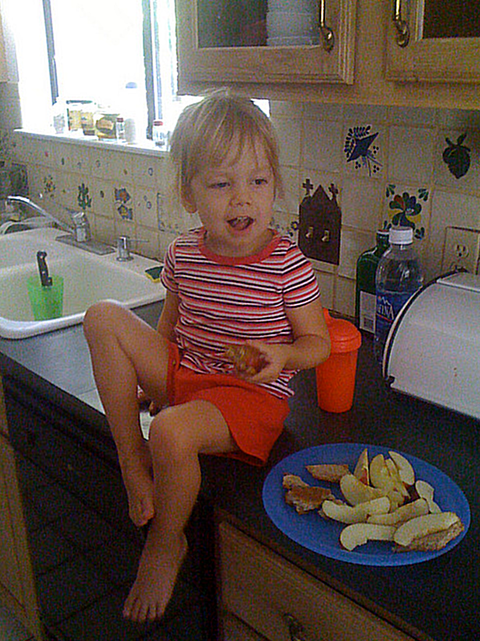}&
  \includegraphics[height=2.73cm]{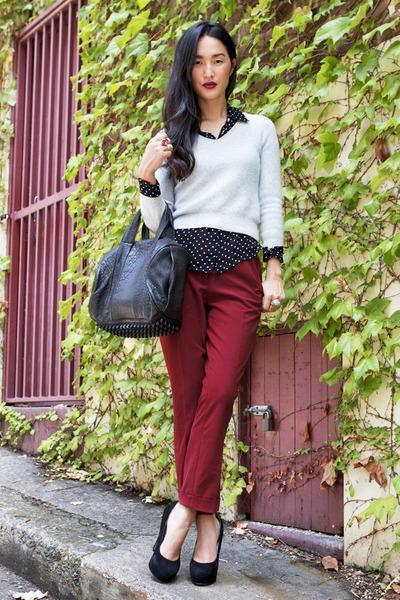}&
  \includegraphics[height=2.73cm]{figures/example_results/kite_base.png}\\
  \includegraphics[height=2.73cm]{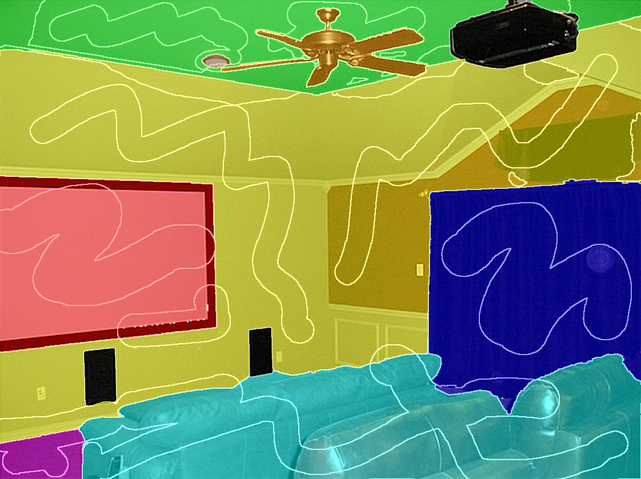}&
  \includegraphics[height=2.73cm]{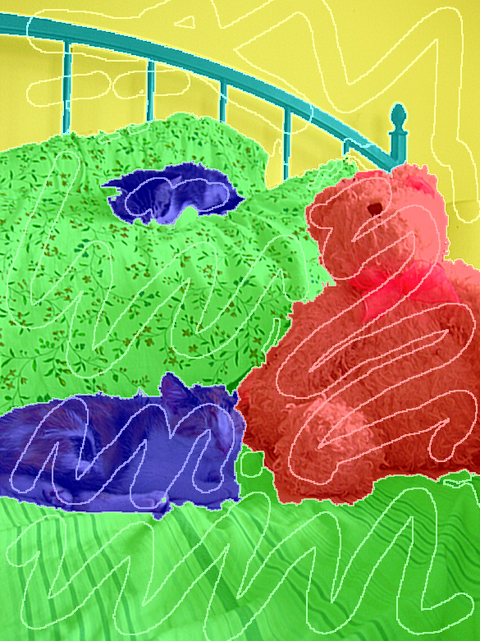}&
  \includegraphics[height=2.73cm]{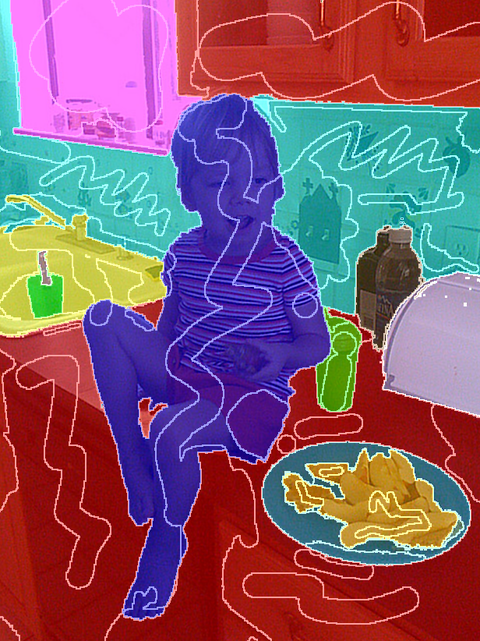}&
  \includegraphics[height=2.73cm]{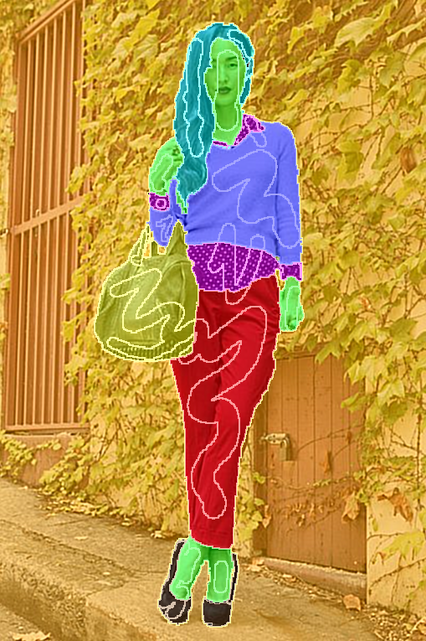}&
  \includegraphics[height=2.73cm]{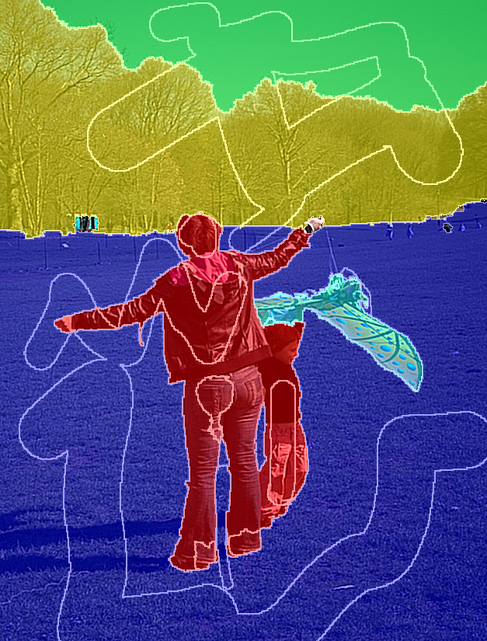}\\
\end{tabular}
\end{center}
\vspace*{-0.5cm}
\caption{A few annotation examples of Magic Paint on ADE-20k (first column), Fashionista (fourth column) and COCO datasets (other columns). The top row shows the original image while the bottom row shows the annotation result, with the drawn strokes. The majority of the strokes are drawn in the center of the objects, while the border strokes are often in freeze foreground mode. Note how typically a large portion of the image is automatically labeled by Magic Paint (all pixels outside any stroke). %
}
\label{fig:example_results}
\end{figure}

%% file: fig_sim_and_human_coco.tex
\begin{figure}[t]
\vspace{-0.2cm}
\begin{center}
 \includegraphics[width=0.48\columnwidth]{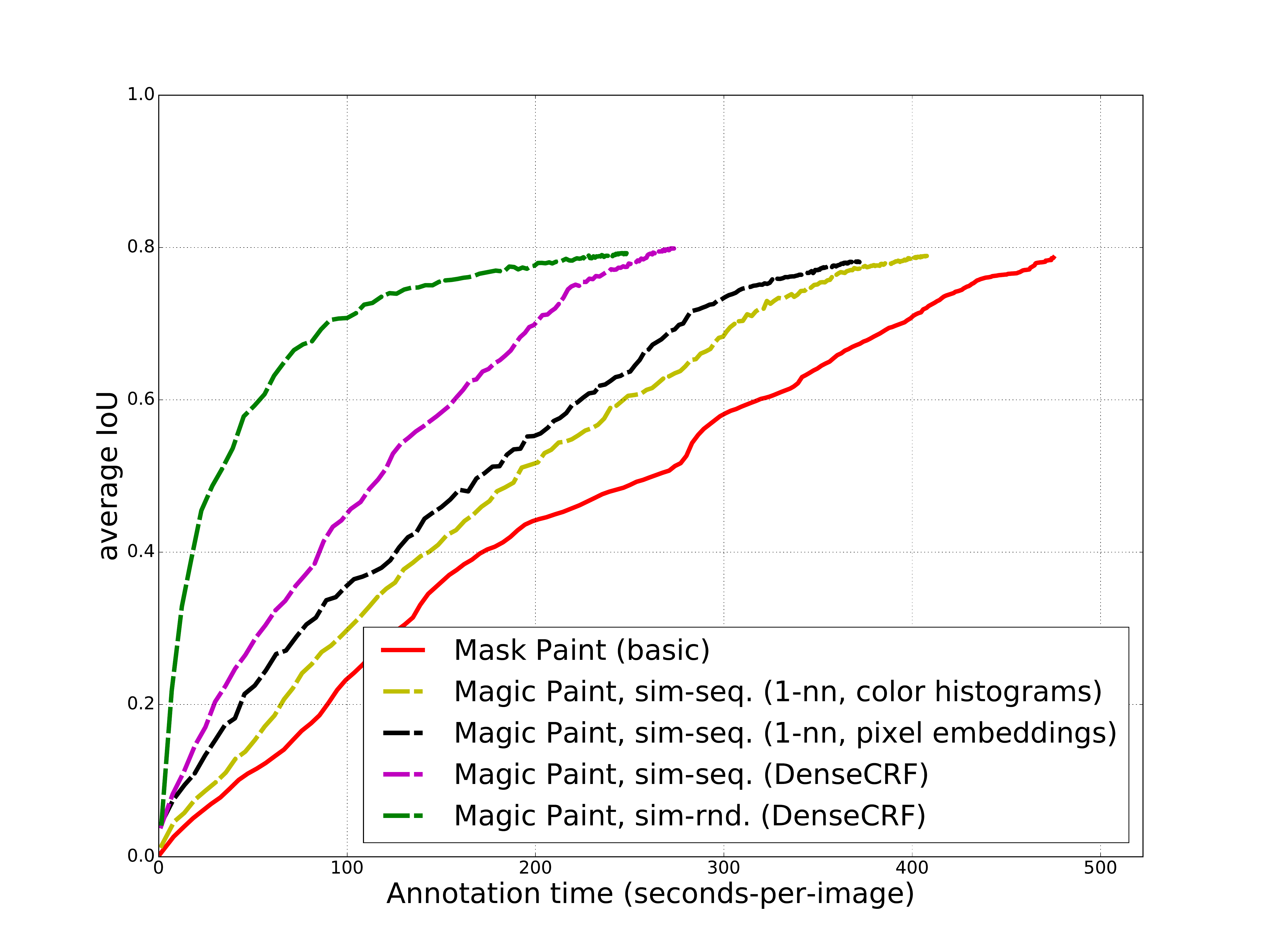} 
\includegraphics[width=0.48\columnwidth]{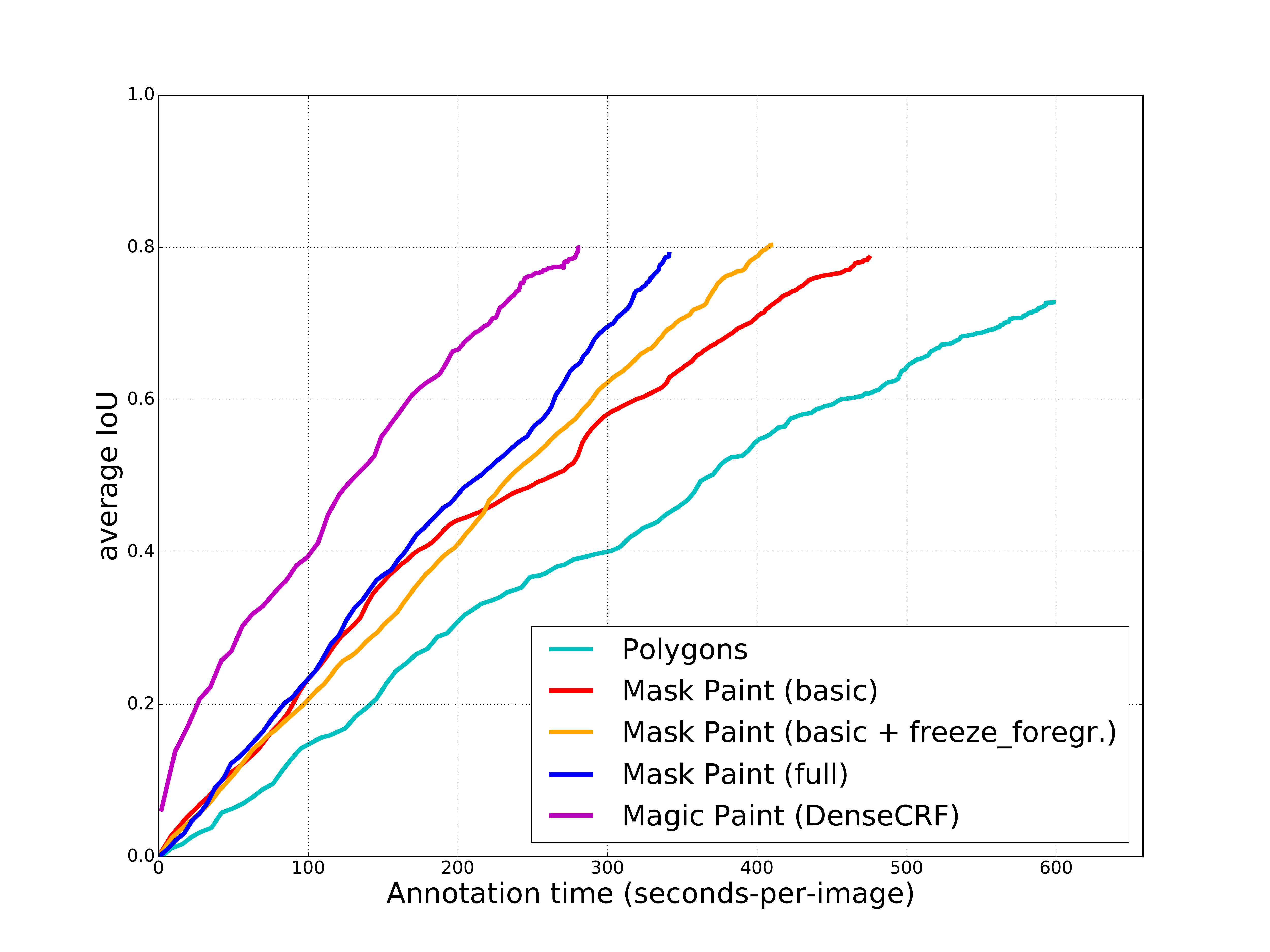}
\end{center}
\vspace*{-0.5cm}
\caption{Experimental results on COCO-val-100. Comparison of model variants in simulation (left) and evaluation with real human annotators (right).}
\label{fig:eval_coco}
\end{figure}

%% file: fig_transfer.tex
\begin{figure}[t]
\vspace{-0.3cm}
\begin{center}
  \includegraphics[width=0.48\linewidth]{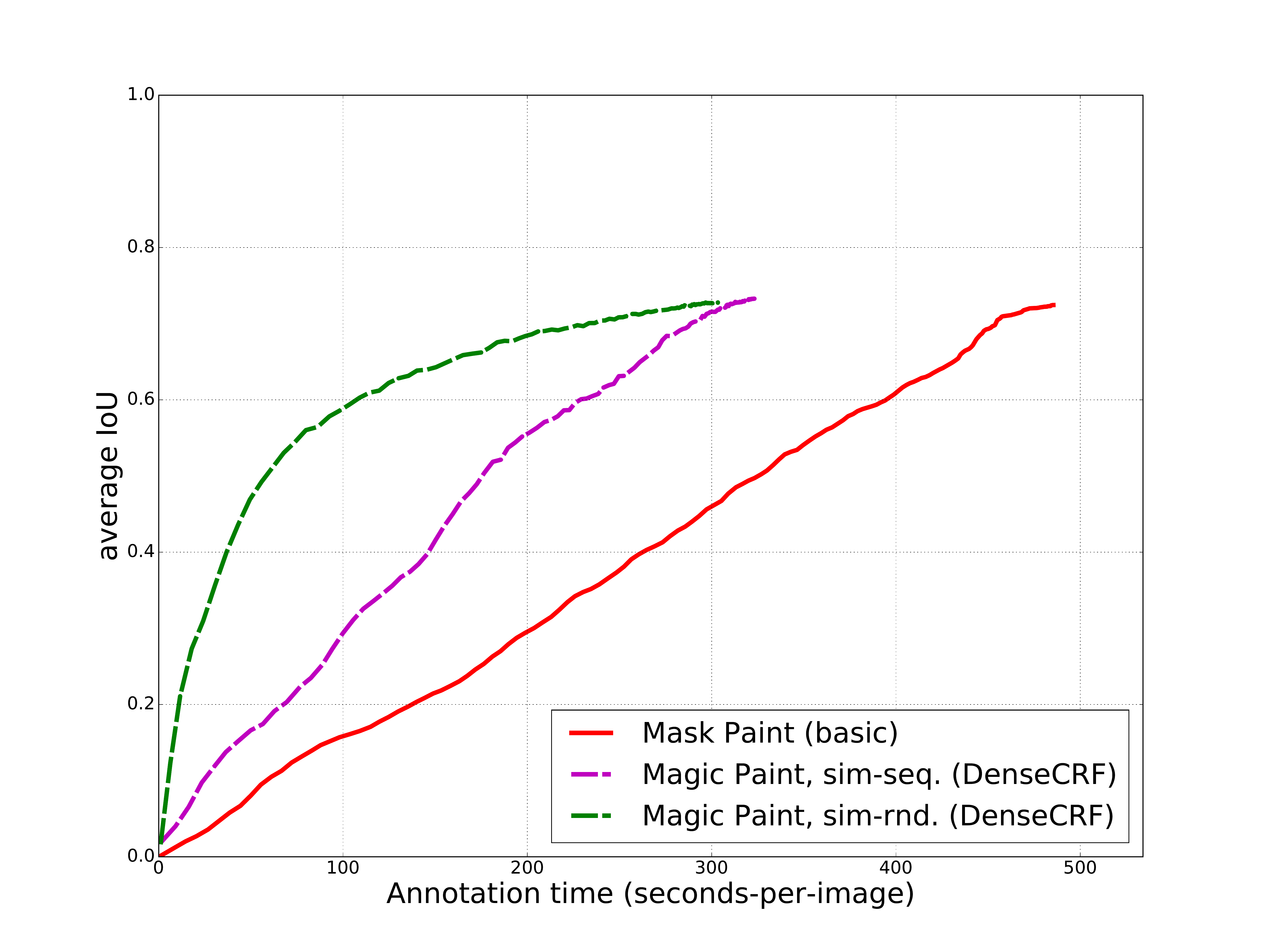}
  \includegraphics[width=0.48\linewidth]{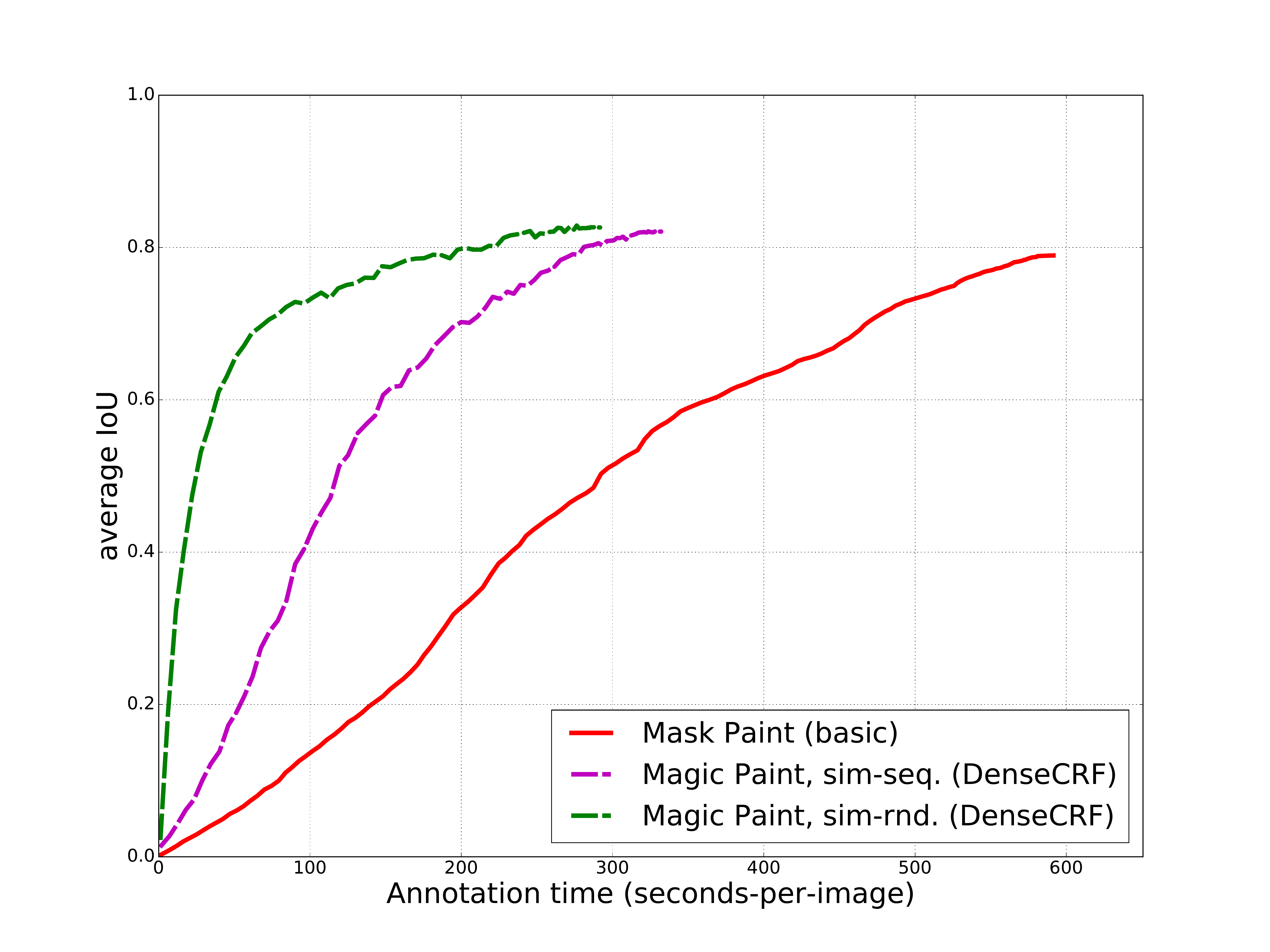}
\end{center}
\vspace{-0.5cm}
   \caption{Evaluation on ADE-20k \cite{zhou17cvpr} and Fashionista \cite{yamaguchi12cvpr} datasets.}
\label{fig:eval_transfer}
\end{figure}

%% file: conclusion.tex
\section{Conclusion}
We presented an annotation tool for collecting training data for full-image semantic segmentation. To enable quick experimentation, we have introduced a novel simulation benchmark. We have experimentally demonstrated that our system reduces the time to annotate full segmentation images by more than a factor $2\times$ compared to traditional polygon drawing techniques.